\documentclass[runningheads]{llncs}

\usepackage{eccv}

\usepackage{eccvabbrv}

\usepackage{marvosym}
\usepackage{graphicx}
\usepackage{booktabs}
\usepackage{algorithm}
\usepackage{algpseudocode}
\usepackage{wrapfig}
\usepackage{amsmath}
\usepackage{bbm}
\usepackage{multirow}
\usepackage{placeins}
\usepackage[accsupp]{axessibility}  

\usepackage{hyperref}

\usepackage{orcidlink}

\begin{document}

\title{InceptionGS: Generative Bootstrapping for Large-Scale Gaussian Splatting under Unstructured View Sampling} 

\titlerunning{InceptionGS}

\author{Tianheng Lu\inst{1}\textsuperscript{$\star$}\orcidlink{0009-0007-7163-5539} \and
Guangyu Wang\inst{1}\textsuperscript{$\star$}\orcidlink{0009-0000-5674-2642} \and
Ruqi Huang\inst{1}\textsuperscript{\Letter}\orcidlink{0000-0001-5942-3671} \and Lu Fang\inst{1}\textsuperscript{\Letter}\orcidlink{0000-0003-3552-0367}}

\authorrunning{T.~Lu et al.}

\institute{\textsuperscript{1}Tsinghua University, Beijing 100084, China}

\maketitle
\footnotetext[1]{\textsuperscript{\Letter}Correspondence to: Lu Fang (fanglu@tsinghua.edu.cn, \href{http://www.luvision.net/}{luvision.net}), Ruqi Huang (ruqihuang@sz.tsinghua.edu.cn, \href{https://rqhuang88.github.io/}{rqhuang88.github.io}).}
\footnotetext[2]{\textsuperscript{$\star$}Authors contributed equally to this work.}

\begin{abstract}
Achieving truly immersive large-scale scene digitization necessitates consistent and visually pleasing rendering across all possible viewing perspectives. However, collecting multi-view images covering every fine detail of a large-scale scene is prohibitive due to scene complexity, capture cost, negligence, or accessibility constraints. As a result, the sampled views tend to be highly \emph{\textbf{unstructured}} -- the majority of the scene is well covered yet certain regions inevitably lack sufficient observations. Existing reconstruction based methods are vulnerable to view scarcity while generation based approaches suffer from generalization, controllability, and 3D consistency issues.
To address this challenge, we propose \emph{\textbf{InceptionGS}}, which bootstraps Gaussian splatting by subtly balancing reconstruction and generation. Starting from an initial Gaussian splatting, InceptionGS reasonably rethinks and repairs problematic regions caused by view scarcity while preserving the quality elsewhere, by softly incorporating scene- and view-adaptive generative priors. Extensive experiments on real-world large-scale scenes demonstrate the superiority and broad applicability of our approach in handling unstructured imagery and boosting high-fidelity Gaussian splatting.
\end{abstract}
\section{Introduction}
\label{sec:intro}
\begin{figure*}[ht]
\begin{center}
    \includegraphics[width=1.0\textwidth]{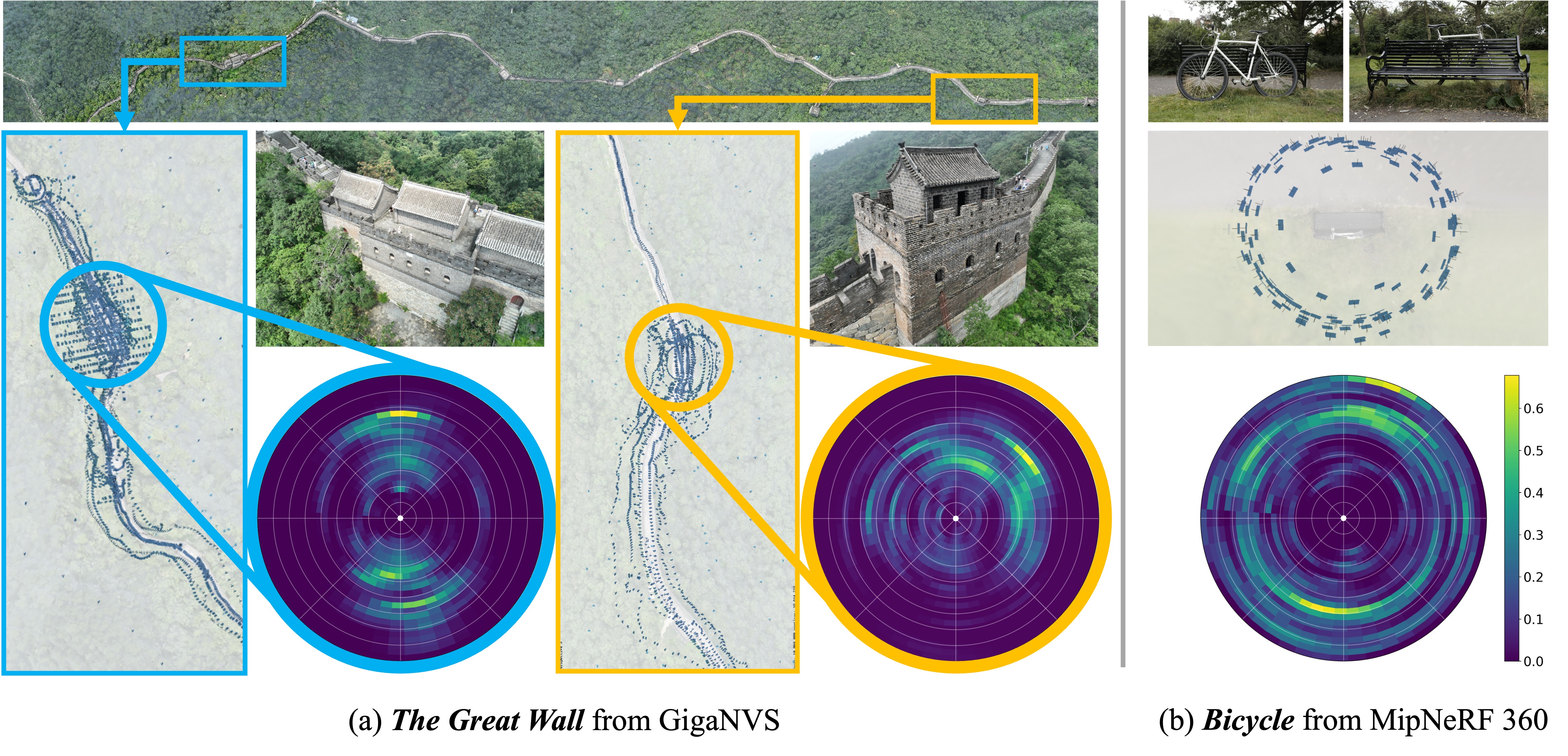}
\end{center}
\vspace{-10pt}
\caption{Comparison of viewpoint distributions. The heatmaps indicate the directional viewing density distributed over the axis-aligned bounding hemisphere of the region of interest, which are approximated by tracing camera rays and accumulating ray-sphere intersections. (a) The unstructured view sampling issue of a real-world large-scale scene from GigaNVS~\cite{wang2024xscale}, with the camera movements (the bird-eye scatter plot) shaped by the complex scene geometry and the viewing density unevenly distributed. (b) A small-scale scene
from MipNeRF360~\cite{barron2022mip} with near-isotropic viewing density.}
\vspace{-25pt}
\label{fig:data}
\end{figure*}
Creating high-quality digital 3D assets of large-scale real-world scenes plays a crucial role in the realms of virtual reality, filmmaking, and gaming. With a one-time multi-view capture, users can effortlessly navigate breathtaking landmarks from the comfort of their home. In such applications, the primary goal is not to reconstruct with absolute precision, but to create controllable, perceptually convincing content grounded in reality. However, delivering truly immersive, lifelike experiences for large-scale scenes remains a hurdle, as it necessitates consistent and visually pleasing rendering across arbitrary viewing directions and scales.
In particular, we identify and address the critical challenges as follows.

First of all, we highlight a practical and pressing issue related to large-scale scene capture, namely \emph{unstructured view sampling}, which stems from 1) \emph{complex camera movements} and 2) \emph{uneven viewing density}. Specifically, unlike small-scale scenes~\cite{barron2022mip, knapitsch2017tanks, ling2024dl3dv, zhou2018stereo} that easily allow for perfectly uniform viewpoint coverage~\cite{kerbl20233d, mildenhall2021nerf, barron2022mip}, the high-quality digitization of large-scale scenes requires highly irregular capturing trajectories shaped by the scene geometry, which oftentimes involves multiple rounds at varying distances for both global structure and fine-grained details. However, due to scene complexity, capture cost, negligence, or accessibility constraints, it is impractical to exhaustively traverse the entire space with all possible view directions. Consequently, the resulting viewpoint density tends to be unevenly distributed -- while most of the scene is sufficiently observed, certain regions inevitably suffer from view scarcity in an uncontrollable manner. In Figure~\ref{fig:data} (a), we visualize the camera distribution of a real-world capture from the GigaNVS dataset~\cite{wang2024xscale}. The directional viewing density of the two selected landmarks reveal a high degree of non-uniformity, especially when compared to the near-isotropic density of a small-scale scene from the MipNeRF360~\cite{barron2022mip} dataset.

\begin{figure*}[t]
\begin{center}
    \includegraphics[width=1.0\textwidth]{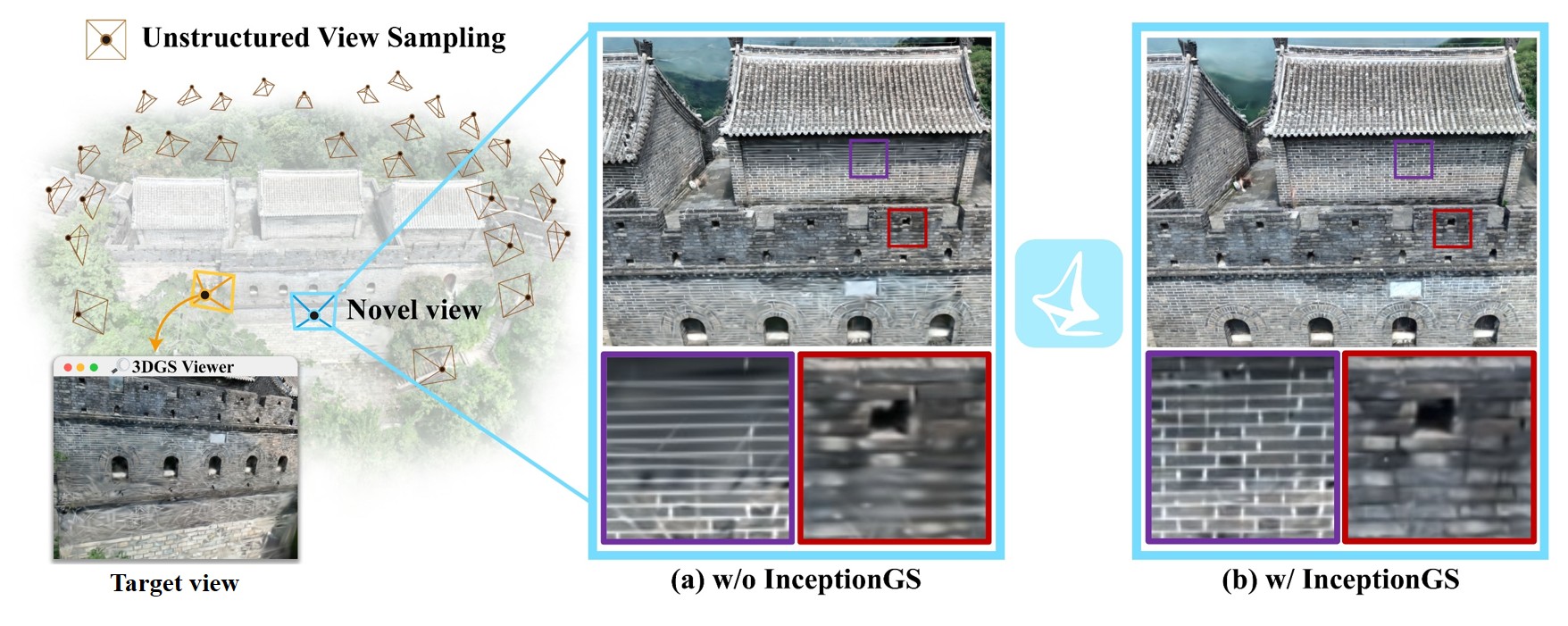}
\end{center}
\vspace{-7pt}
\caption{We propose \emph{InceptionGS} to address the \emph{unstructured view sampling} challenge related to real-world large-scale scene capture. Starting from an initial Gaussian splatting and given a target viewpoint indicating visual artifacts, InceptionGS reasonably rethinks and refines the scene by softly incorporating scene-adaptive generative priors in a view-adaptive manner, enabling robust, high-fidelity Gaussian splatting in the wild.
}
\vspace{-20pt}
\label{fig:teaser}
\end{figure*}

Unfortunately, existing reconstruction based and generation based approaches struggle to address this challenge. Reconstruction based novel view synthesis (NVS) methods, such as 3D Gaussian splatting (3DGS)~\cite{kerbl20233d} and its extensions~\cite{chen2024pgsr, yu2024mip, kheradmand20243d, park2025dropgaussian, chen2024splatformer}, optimize the explicit primitive-based 3D representation in a scene-specific manner. Despite the remarkable photo-realism they achieve for in-the-wild scenes, the reconstruction quality drastically deteriorates under large variation from training views, which hinders the sense of immersion, as shown in Figure~\ref{fig:teaser} (a). Although the most effective way to fix these artifacts is additional data collection, it is laborious and often impractical for real-world applications.

On the other end of the spectrum, recent advances in generation based NVS~\cite{you2024nvs, chen2025mvsplat360, yu2024viewcrafter, liu2024novel, liu20253dgs, he2024cameractrl, liang2024wonderland} have demonstrated compelling results using one or few input views, leveraging strong priors from video~\cite{xing2024dynamicrafter, blattmann2023stable, yang2024cogvideox} or multi-view~\cite{wu2024reconfusion, gao2025cat3d, zhou2025stable} diffusion models. They retarget NVS as a feed-forward generation process, leveraging training-free guidance~\cite{you2024nvs, liu2024novel} or conditional finetuning~\cite{bahmani2024ac3d, yu2024viewcrafter, chen2025mvsplat360, liu20253dgs, wu2025difix3d+}. However, these methods have difficulty generalizing to complex large-scale scenes, often leading to structural distortions and appearance hallucinations due to the scarcity of high-quality training data. 
Besides, these approaches condition the generation either on Plücker embedding~\cite{he2024cameractrl, liang2024wonderland} or artifact-prone RGB rendering~\cite{you2024nvs, chen2025mvsplat360, yu2024viewcrafter, liu2024novel, liu20253dgs, wu2025difix3d+}, which are hard to ensure precise control and long-term 3D consistency. Therefore, synthesizing complex large-scale scenes purely from a generative perspective remains a non-trivial obstacle.

In light of the above observations, 
rather than focusing solely on reconstruction or generation, we delve into the intersection of both paradigms, aiming to subtly integrate the strengths of these two extremes to push the boundaries of immersive large-scale scene digitization. Our key insight is that reconstruction could provide meaningful scene-specific clues to customize generation, the adapted generation, in turn, effectively boosts reconstruction by characterizing the underlying statistics. To this end, we introduce \emph{InceptionGS}, a novel bootstrapping approach exploiting this synergy -- with reconstruction and generation mutually ``incepted'' into each other -- to address the unstructured sampling issue of large-scale scenes, as shown in Figure~\ref{fig:teaser}. Specifically, we propose to adapt the pretrained generative prior with scene-specific geometric guidance from an initial 3DGS reconstruction. 
Then, InceptionGS reasonably rethinks and refines the 3DGS field by injecting the scene-adaptive generative prior on a set of virtual viewpoints indicating visual artifacts obtained through view selection and interpolation. This ensures a long-range refinement in 3D space and avoids overfitting to the specified view. In contrast to existing approaches~\cite{yang2024gaussianobject, liu20253dgs, chen2025mvsplat360, yu2024viewcrafter, liu2024novel, you2024nvs} that conduct generative refinement upon low-quality, artifact-prone RGB renderings, InceptionGS directly learns the scene-specific appearance distribution conditioned on the geometry. This design choice stems from the observation that geometry is strictly 3D consistent while being more robust than appearance given the well-behaved planar regularization~\cite{furukawa2009accurate, schonberger2016pixelwise, chen2024pgsr}. Moreover, the underlying geometry-appearance correspondence exhibits a strong inductive bias in terms of local recurrence, which can be effectively characterized by convolutional kernels to facilitate generative learning.


Unlike prior works~\cite{wu2024reconfusion, gao2025cat3d, yu2024viewcrafter, liu2024novel, liang2024wonderland} that directly distill the generative prior into a 3D representation, our approach adopts a fundamentally different strategy by \emph{softly} incorporating the prior to better represent reconstructive details. Specifically, we blend the generated images with reliable photometric clues~\cite{buehler2001unstructured, rong2022boosting} and propose a view-space importance sampling scheme to adaptively focus on the most informative views for bootstrapping, thus mitigating the conflicts between reconstruction and generation. Notably, InceptionGS effectively bootstraps 3DGS field without requiring additional resource-intensive real-world data collection, and the refinement can be iterated to repair problematic regions incrementally while preserving the quality of previously well-reconstructed areas.

To demonstrate the efficacy of InceptionGS, we simulate unstructured view sampling by clustering and filtering real-world imagery and establish a novel benchmark primarily based on the challenging GigaNVS dataset~\cite{wang2024xscale}. Through extensive quantitative and qualitative evaluations, we demonstrate the state-of-the-art performance of InceptionGS in handling unstructured sampling, which consistently improves reconstruction fidelity and visual coherence across diverse large-scale scenes. Remarkably, our method effectively reduces the average FID by 32\% after bootstrapping and significantly outperforms the top-performing diffusion based alternative by {10\%} in FID, and {14\%} in LPIPS, underscoring a substantial advancement towards truly immersive large-scale scene digitization. In summary, our main contributions are as follows: 
\begin{itemize}
    \item We identify a practical unstructured view sampling challenge related to real-world large-scale scenes, for which we establish a benchmark to support comprehensive evaluations.
    \item We introduce InceptionGS to bootstrap large-scale 3DGS by adapting the generic generative prior with on-the-fly, hierarchical geometric signals as conditioning, subtly integrating the strengths of reconstruction and generation. 
    \item We propose photometric blending and view-adaptive sampling to softly inject the adapted generative prior, effectively preserving fine details while mitigating the conflicts between reconstruction and generation.
    \item We demonstrate significant improvements over reconstruction based and generation based alternatives, both quantitatively and qualitatively.  
\end{itemize}

\section{Related Work}
\label{sec:related_work}


\noindent{\textbf{Reconstruction based NVS.}} Per-scene optimization approaches~\cite{mildenhall2021nerf, kerbl20233d, muller2022instant, wang2024xscale, fridovich2022plenoxels} encapsulate the scene-specific information using implicit neural networks~\cite{mildenhall2021nerf}, explicit feature structures~\cite{kerbl20233d, fridovich2022plenoxels} or hybrid representations~\cite{muller2022instant, wang2024xscale}, enabling photorealistic rendering given perfect viewpoint coverage. By pretraining deep neural networks on diverse multi-view data~\cite{barron2022mip, knapitsch2017tanks, ling2024dl3dv, zhou2018stereo, liu2021infinite}, generalizable NVS~\cite{chen2024splatformer, yu2021pixelnerf, chen2024mvsplat, riegler2021stable, flynn2019deepview} bypasses the costly optimization process and synthesizes novel views from few input views in a feed-forward manner. 
However, the reconstruction quality drastically deteriorates under large variations in viewpoint or scale, exhibiting excessive blurries and needle-like artifacts once deviating from the training views.

\noindent{\textbf{Generation based NVS. }} Benefiting from the rapid advances in diffusion models~\cite{yang2024cogvideox, kong2024hunyuanvideo, blattmann2023stable, rombach2022high, zhang2023adding, xing2024dynamicrafter}, generative NVS~\cite{wu2024reconfusion, gao2025cat3d, liang2024wonderland, you2024nvs, chen2025mvsplat360, yu2024viewcrafter, liu2024novel, liu20253dgs, bahmani2024ac3d, wu2025difix3d+, zhou2025stable} approaches have demonstrated strong capabilities in completing unseen regions with plausible content, enabling both object-level~\cite{gao2025cat3d} and scene-level~\cite{liang2024wonderland, yu2024viewcrafter, you2024nvs} generation, even from a single image. A notable trend in recent works bridges video generation with NVS by applying temporal restoration on an initial low-quality estimate obtained via image warping~\cite{you2024nvs}, point-cloud rendering~\cite{yu2024viewcrafter}, or few-view 3DGS reconstruction~\cite{chen2025mvsplat360, liu20253dgs, liu2024novel}. Nevertheless, these methods often lead to hallucinations, struggling to generalize to complex large-scale scenes that rarely occur in the training data. Although the rich Internet collections~\cite{deitke2023objaverse1, deitke2023objaverse, ling2024dl3dv, chen2024panda, zhou2018stereo, liu2021infinite} can well support the generation of common small-scale scenes~\cite{zhou2018stereo} with simple camera movements, they fail to fully characterize the diversity, complexity, and uniqueness of real-world sceneries~\cite{wang2024xscale} and the scene-dependent capturing trajectories. On the other hand, even the state-of-the-art diffusion based scene-repairing model~\cite{wu2025difix3d+} trained on millions of data have difficulty understanding the ever-extending structure of the Great Wall, or the intricate geometry of lintels and cornices in traditional Chinese architectures. Also, it lacks precise control and fails to strictly maintain 3D consistency.
\section{Methodology}
\label{sec:method}
In this section, we present InceptionGS, which aims to address the challenges raised by the unstructured view sampling of large-scale scene capture. Given a set of calibrated multi-view images $\{\mathcal{I}_i \in \mathbb{R}^{H\times W\times 3}\}_{i=1}^{N}$, InceptionGS first optimizes a 3DGS field $\{\mathcal{G}_{i}\}$ for initialization while simultaneously adapts the generic generative prior with scene-specific geometric features. 
InceptionGS then refines the 3DGS field by softly incorporating the customized generative prior with reliable photometric clues in a view-adaptive manner, effectively repairing the 3DGS field from a set of candidate virtual viewpoints generated via view selection and interpolation. This process can be iteratively conducted, enabling a self-evolving 3DGS that robustly handles unstructured sampling, without requiring additional real-world data collection. An illustration of the InceptionGS pipeline is shown in Figure~\ref{fig:pipeline}, and a formal description is outlined in the Appendix.

In the following, we first briefly review the vanilla 3DGS and PGSR approach (Section \ref{subsec:prelim}). We then detail our design on individual core components, including generative adaptation (Section \ref{subsec:scene_adap}), photometric blending (Section \ref{subsubsec:warp}), and the view sampling scheme (Section \ref{subsubsec:LMC}). 

\begin{figure*}[htbp]
    \vspace{-10pt}
    \centering
    \newcommand{\colw}{0.19}
    \newcommand{\figw}{1} 
    \includegraphics[width=\figw\textwidth,trim={0cm 0cm 0cm 0cm},clip]{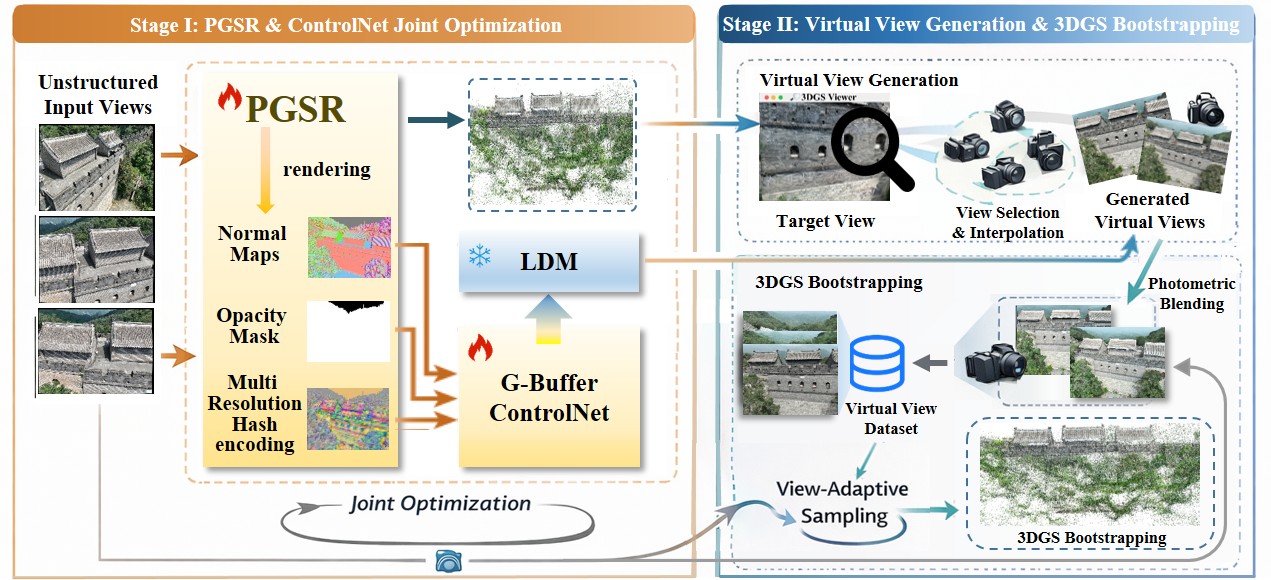}
    \hfill
\vspace{-6pt}
\caption{Overview of the generative bootstrapping pipeline with InceptionGS. Stage I: We perform generative adaptation (Section \ref{subsec:scene_adap}) by adapting the generative prior of a pretrained LDM $\epsilon_{\theta}(\cdot)$ through G-buffer conditioning, with the guidance of an initial 3DGS optimization. Stage II: Given a target viewpoint $j^{\star}$ exhibiting visual artifacts, we generate a set of candidate virtual viewpoints $V_{\text{vrt}}$ for generative bootstrapping, using view selection and interpolation. During bootstrapping, we enhance the reconstructive details by blending the generative prior with reliable photometric clues (Section \ref{subsubsec:warp}) and we softly inject generative supervision through view-space importance sampling (Section \ref{subsubsec:LMC}).}
\vspace{-25pt}
\label{fig:pipeline}
\end{figure*}

\subsection{Preliminaries}
\label{subsec:prelim}
3DGS~\cite{kerbl20233d} explicitly represents the scene as a collection of 3D Gaussian primitives $\{\mathcal{G}_{i}\}$, where each primitive $\mathcal{G}_{i}$ is parameterized as a 3D Gaussian ellipsoid with centroid $\boldsymbol{\mu}_{i} \in \mathbb{R}^3$ and covariance matrix $\boldsymbol{\Sigma}_{i} = \boldsymbol{R}_i\boldsymbol{S}_i\boldsymbol{S}^{{T}}_{i}\boldsymbol{R}_{i}^{{T}}$:
\begin{equation}\label{eqn:3dgs}
    \mathcal{G}_{i}(\boldsymbol{x} | \boldsymbol{\mu}_{i}, \boldsymbol{\Sigma}_{i}) = e^{-\frac{1}{2}(\boldsymbol{x} - \boldsymbol{\mu}_{i})^{{T}}\boldsymbol{\Sigma}^{-1}_{i}(\boldsymbol{x} - \boldsymbol{\mu}_{i})}.
\end{equation}

To render the RGB image $\mathcal{C} \in \mathbb{R}^{H\times W\times 3}$ from a given viewpoint, 3DGS alpha-blends the color $\boldsymbol{c}_{i}$ contributed by each Gaussian primitive $\mathcal{G}_{i}$ on the ray, in a fast tile-based manner:
\begin{equation}\label{eqn:3dgs_color_render}
    {\mathcal{{C}}} = \sum_{i}{T_{i}\alpha_{i}\boldsymbol{c}_{i}},
\end{equation}
where $\alpha_{i}$ is the opacity value contributed by the respective Gaussian primitive $\mathcal{G}_{i}$, and $T_{i}$ is the cumulative opacity.

The 3DGS field $\{\mathcal{G}_{i}\}$ is optimized through differentiable rendering by minimizing a combination of L1 and SSIM loss: 
\begin{equation}\label{eqn:3dgs_loss}
    \mathcal{L}_{C} = \lambda_1 \mathcal{L}_{1} + (1 - \lambda_1)\mathcal{L}_{\text{D-SSIM}}.
\end{equation}

To enhance the underlying geometry quality of 3DGS, PGSR~\cite{chen2024pgsr} draws inspiration from traditional patch-matching MVS~\cite{furukawa2009accurate, schonberger2016pixelwise} and incorporates local planar priors to enable accurate and smooth surface reconstruction. Specifically, it flattens each Gaussian ellipsoid into local planes and applies a variety of auxiliary planar regularizations:
\begin{equation}\label{eqn:pgsr_loss}
    \mathcal{L}_{G} = \lambda_{SV} \mathcal{L}_{SV} + \lambda_{MVC} \mathcal{L}_{MVC} + \lambda_{MVG} \mathcal{L}_{MVG},
\end{equation}
where $\mathcal{L}_{SV}$ denotes the single-view normal smoothness term, which regularizes the alpha-blended planar normal $\mathcal{N}=\sum_{i}{T_{i}\alpha_{i}\boldsymbol{n}_{i}}$ to align with that derived from the local depth gradients. $\mathcal{L}_{MVC}$ denotes the NCC-based photo-consistency constraint~\cite{furukawa2009accurate, schonberger2016pixelwise}, and $\mathcal{L}_{MVG}$ is the geometric consistency term penalizing the reprojection error~\cite{schonberger2016pixelwise, yao2018mvsnet}.

\subsection{Stage I: generative adaptation}
\label{subsec:scene_adap}

 We observe that geometry can be more effectively regularized than appearance due to its inherent 3D consistency and the local planar nature~\cite{chen2024pgsr, schonberger2016pixelwise, furukawa2009accurate}, which makes it a reliable condition for appearance generation. Besides, for a 3D scene, there exists intrinsic self-similarity~\cite{hanocka2020point2mesh} regarding the local correspondence between geometry and appearance. Thus, we can exploit geometric clues to ease the generative learning of the appearance distribution, with shared convolutional kernels applied on surface patches to characterize the internal reoccurring mapping modes. To this end, we repurpose a pretrained text-to-image latent diffusion model (LDM)~\cite{rombach2022high} to learn the scene-specific statistics of the geometry-appearance correspondence by conditioning it on an additional geometry buffer and conducting data-efficient finetuning via ControlNet~\cite{zhang2023adding}. 

\noindent{\textbf{Geometry conditioning.}} We build upon PGSR~\cite{chen2024pgsr} due to its superior geometry quality and unbiased rendering. We rasterize the planar normal map $\mathcal{N} \in \mathbb{R}^{H\times W\times 3}$ and the foreground opacity mask $\mathcal{O} \in \mathbb{R}^{H\times W\times 1}$ as geometry buffer (G-buffer) to condition appearance generation. To further strengthen controllability and 3D consistency, we introduce a learnable, surface-aware, multi-resolution hash feature map $\mathcal{F} \in \mathbb{R}^{H\times W\times Z}$~\cite{wang2024xscale, muller2022instant} as additional conditioning. The input for the ControlNet is the concatenation of the rasterized buffer $\mathcal{R} = (\mathcal{N}, \mathcal{O}, \mathcal{F}) \in \mathbb{R}^{H\times W\times (3+1+Z)}$, which is computed on-the-fly from the continuously updating GS model at each gradient descent step, so as to ensure the robustness against varying geometry quality. Please refer to the Appendix for more implementation details.

\noindent{\textbf{Adapting generative prior.}} To better leverage the self-similarity of the geometry-appearance correspondence, we employ the G-buffer conditional generative model on local image patches $\hat{\mathcal{I}} \in \mathbb{R}^{s\times s\times 3}$, which are randomly sampled from the multi-view images $\{\mathcal{I}_i \in \mathbb{R}^{H\times W\times 3}\}_{i=1}^{N}$ at each training step. Let $\epsilon_{\theta}(\cdot)$ denote the denoising diffusion UNet with learnable ControlNet parameters $\theta$, the diffusion loss can be formulated as:
\begin{equation}\label{eqn:diffusion_loss}
    \mathcal{L}_{D} =  \mathbb{E}_{t \sim {U}(0,1),\epsilon \sim {N}(\boldsymbol{0}, \boldsymbol{1})}[\|\epsilon_{\theta}(\boldsymbol{z}_{t}, t, \hat{\mathcal{R}}) - \epsilon\|^2_2],
\end{equation}
where $\epsilon$ is the sampled random noise, $\hat{\mathcal{R}}\in \mathbb{R}^{s\times s\times (3+1+Z)}$ is the local patch of the G-buffer $\mathcal{R}$ corresponding to $\hat{\mathcal{I}}$. We denote by $\boldsymbol{z}_t =\sqrt{\alpha_t} \boldsymbol{z}_0 + \sqrt{1-\alpha_t} \epsilon$ the noisy latent at timestep $t$, with $\boldsymbol{z}_0 = \mathcal{E}(\hat{\mathcal{I}})$ being the VAE latent of the corresponding image patch $\hat{\mathcal{I}}$. 

During this stage, we jointly optimize the ControlNet and the 3DGS field at the same time, using a combination of loss functions including the RGB loss $\mathcal{L}_{C}$ (Eq.~\ref{eqn:3dgs_loss}), the geometry regularization loss $\mathcal{L}_{G}$ (Eq.~\ref{eqn:pgsr_loss}), and the diffusion loss $\mathcal{L}_{D}$ (Eq.~\ref{eqn:diffusion_loss}).
We use DDIM scheduling in the denoising process and use MultiDiffusion~\cite{bar2023multidiffusion} to synthesize the full-resolution image by fusing diffusion trajectories of the sliding patches. We denote by ${\boldsymbol{z}}^{\ast}_{0} \sim \text{DDIM}(\boldsymbol{z}_T, \epsilon_{\theta}(\cdot, \cdot, {\mathcal{R}}))$ the final denoised latent and by $\mathcal{I}^{\ast} = \mathcal{D}({\boldsymbol{z}}^{\ast}_{0})$ the generated image from the VAE decoder.

As will be demonstrated in Section \ref{subsec:benchmark} and Section \ref{subsec:ablation}, the proposed scene-adaptive generative model substantially improves the quality and expressivity of appearance generation for fine details, while simultaneously enabling remarkable 3D consistency and controllability without any reliance on temporal modelling, compared to other generative methods~\cite{yu2024viewcrafter, you2024nvs, bahmani2024ac3d, wu2025difix3d+}. 
\subsection{Stage II: bootstrapping Gaussians}
So far we have proposed an effective approach to adapt the LDM $\epsilon_{\theta}(\cdot)$ for characterizing the scene-specific appearance distribution conditioned on the reconstructed geometry. Then, a straightforward way to enhance the 3DGS field would be sampling a set of unseen viewpoints ${V}_{\text{vrt}}$ and supervising the RGB rendering $\mathcal{C}$ with the generated appearance $\mathcal{I}^{\ast}$.

\begin{figure*}[htbp]
    \vspace{-6pt}
    \centering
    \newcommand{\colw}{0.19}
    \newcommand{\figw}{1} 
    \includegraphics[width=\figw\textwidth,trim={0cm 0cm 0cm 0cm},clip]{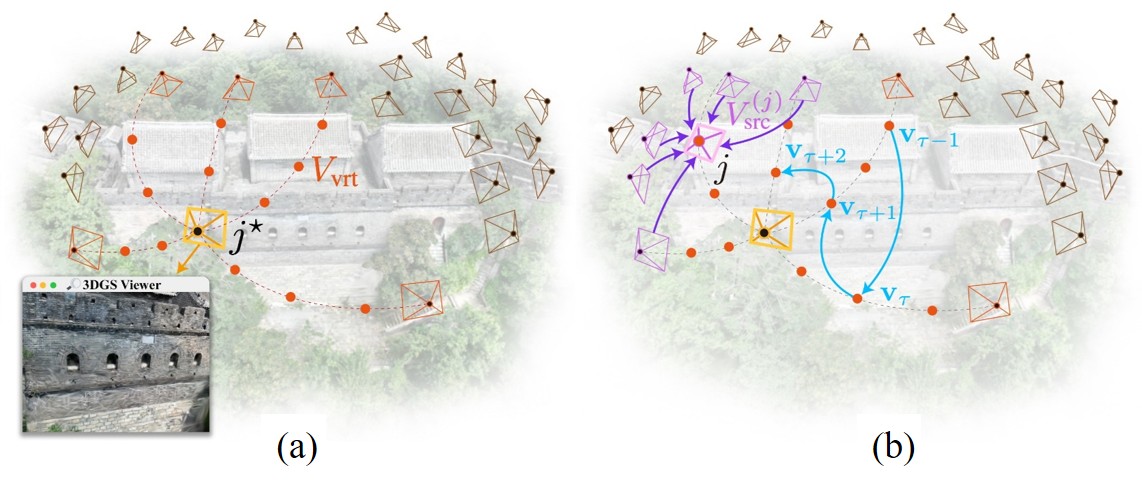}
    \hfill
\vspace{-16pt}
\caption{Illustration of 3DGS bootstrapping. (a): Given a target view $j^{\star}$, we generate a set of candidate virtual viewpoints $V_{\text{vrt}}$ through principled view selection and interpolation. (b): During the bootstrapping stage, reconstructive fidelity is improved by integrating generative priors with reliable photometric evidence. Meanwhile, generative supervision is softly incorporated through view-space importance sampling, enabling stable and geometry-consistent refinement.}
\label{fig:PB&VS}
\end{figure*}

\noindent{\textbf{Virtual viewpoint generation.}} 
As shown in Figure \ref{fig:PB&VS} (a), let ${j}^{\star}$ denote a target viewpoint indicating erroneous visual artifacts, we sample a set of candidate virtual viewpoints ${V}_{\text{vrt}}$ by firstly selecting the top-$K$ nearest input training views ${V}_{\text{src}}^{({j}^{\star})}$~\cite{yao2018mvsnet,schonberger2016pixelwise}, considering both camera position and orientation, and then linearly interpolate the camera center and the quaternion~\cite{shoemake1985animating} between ${j}^{\star}$ and each selected training view $i \in {V}_{\text{src}}^{({j}^{\star})}$:
\begin{equation}\label{eqn:view_interp}
    {V}_{\text{vrt}} = \{\text{interp}({j}^{\star}, i, l) | i \in {V}_{\text{src}}^{({j}^{\star})}, l \in [1, L]\},
\end{equation}
where $l$ is the index of the interpolation, and we sample a total of $|{V}_{\text{vrt}}|=LK$ candidate virtual viewpoints.

\noindent{\textbf{Photometric blending.}}\label{subsubsec:warp}  In practice, we find that the naive approach -- namely, directly using the generated appearance $\mathcal{I}^{\ast}$ at virtual viewpoints ${V}_{\text{vrt}}$ as pseudo training views -- already demonstrates effectiveness in tackling the unstructured view sampling challenge for large-scale scenes. Nevertheless, a remaining issue is that the generated images still have certain domain shifts from the real-captured ones, and the conditional LDM fails to ensure the photometric consistency of high-frequency details encapsulated in the VAE.

To take a step further, 
we propose to enhance the textural details using photometric clues from the real-captured training views~\cite{buehler2001unstructured}, which is illustrated as the purple arrows in Figure \ref{fig:PB&VS} (b). 
Specifically, for each sampled unseen viewpoint $j \in {V}_{\text{vrt}}$, we select the top-$K$ nearest training views ${V}_{\text{src}}^{(j)}$ similar to~\cite{schonberger2016pixelwise, yao2018mvsnet}. We then warp the RGB images $\{\mathcal{I}_i| i \in {V}_{\text{src}}^{(j)}\}$ of the selected training views into the viewpoint $j$, using the camera intrinsics and extrinsics, and the depth renderings of the initial PGSR reconstruction. We denote by $\{\mathcal{I}_{i \rightarrow j}| i \in {V}_{\text{src}}^{(j)}\}$ the set of warped RGB images at the unseen viewpoint $j$, and by $\{{W}_{i \rightarrow j}| i \in {V}_{\text{src}}^{(j)}\}$ the corresponding blending weights measuring visibility~\cite{rong2022boosting}, which are modulated by a threshold parameter $\delta$ to suppress contributions from potentially occluded regions. Leveraging photometric consistency, we perform image-based rendering by a weighted average of the warpings:
\begin{equation}\label{eqn:warp}
    \widetilde{\mathcal{I}}_j=\sum_{i\in {V}^{(j)}_{\text{src}}}W_{i\rightarrow j} \mathcal{I}_{i \rightarrow j}.
\end{equation}

The image-based rendering $\widetilde{\mathcal{I}}_j$ in Eq.~\ref{eqn:warp} inherits high-quality textural details from the real-world captures when the geometry is well-approximated. However, it is vulnerable to occlusion, intricate geometry, and large viewpoint variation. To address this issue, we set $\delta$ in a conservative way to ensure that only trustworthy evidence from source views is retained in the fusion, and then blend the generated image $\mathcal{I}^{\ast}_{j}$ only with the most reliable photometric clues:
\begin{equation}\label{eqn:blend}
    \mathcal{I}^{\ast}_{j} \leftarrow M \circ \widetilde{\mathcal{I}}_j + (\mathbf{1}-M) \circ \mathcal{I}^{\ast}_{j},
\end{equation}
where $M$ is the binary mask that guides the completion of missing regions with the generative prior. More details about photometric blending can be found in the Appendix.

To refine the 3DGS field, we apply a combination of structural and perceptual loss on each sampled virtual view $j \in {V}_{\text{vrt}}$:
\begin{equation}\label{eqn:photo_loss}
    \begin{cases}
    \mathcal{L}_{P}^{(j)} =\lambda_2  \mathcal{L}_{\text{LPIPS}}(\mathcal{C}_j, \mathcal{I}^{\ast}_j),  ~~~~~~~~~~~~~~~\text{if} \ \ \mathcal{L}_{\text{SSIM}}(\mathcal{C}_j, \mathcal{I}^{\ast}_j) \leq \tau,  \\
    \mathcal{L}_{P}^{(j)} =  \lambda_3 \mathcal{L}_{\text{1}}(\mathcal{C}_j, \mathcal{I}^{\ast}_j) + \lambda_4 \mathcal{L}_{\text{SSIM}}(\mathcal{C}_j, \mathcal{I}^{\ast}_j),  ~~~~~~~~\text{otherwise}.
    \end{cases}
\end{equation}
We employ reconstructive supervision in a conservative manner, and once the structural loss $\mathcal{L}_{\text{SSIM}}$ is below a predefined threshold $\tau$, we resort to perceptual supervision $\mathcal{L}_{\text{LPIPS}}$.

\noindent{\textbf{View-adaptive sampling.}} \label{subsubsec:LMC} Equipped with the scene-adaptive generative prior and the photometric blending mechanism, InceptionGS could effectively refine the initial 3DGS field at unseen virtual viewpoints. The remaining issue is how we can efficiently sample the set of unseen viewpoints for bootstrapping. 

 Instead of uniformly injecting the generative prior across ${V}_{\text{vrt}}$, we propose to sample the most informative virtual view adaptively at each training step based on Langevin Monte Carlo (LMC) importance sampling~\cite{brosse2018promises, kheradmand2024accelerating}. Intuitively, we hope to strike a better balance between generative appearance regularization and photometric 3DGS reconstruction, thus preventing the over-exposure of the generative supervision from degrading photometric optimization. 
 
\begin{wrapfigure}{r}{0.5\textwidth}
\begin{minipage}{0.5\textwidth}
\vspace{-36pt}
\begin{algorithm}[H]
\scriptsize
\caption{View-adaptive sampling $\mathcal{S}$.}\label{algo:mcmc}
\begin{algorithmic}[1]
\Require Initial viewpoint $q$, 3DGS field $\{\mathcal{G}_i\}$, candidate viewpoint set ${V}_{\text{vrt}}$, pose vector set $\{\boldsymbol{\mathrm{v}}_j | j \in {V}_{\text{vrt}}\}$, blended images $\{\mathcal{I}^{\ast}_j | j \in {V}_{\text{vrt}}\}$, view sampling step $T_{\text{vs}}$

\State Initialize $\boldsymbol{\mathrm{v}}_0 = \boldsymbol{\mathrm{v}}_q$
\For{$\tau = 0, \ldots, T_{\text{vs}}$} \Comment{Iterate over sampling steps}
    \State $\mathcal{P} = \exp(\|\mathcal{C}_q - \mathcal{I}^{\ast}_q\|_1)$ \Comment{Get distribution}
    \State $\mathcal{G}_i = \mathcal{G}_i-\lambda_{\text{lr}}\nabla_{\mathcal{G}_i} (\mathcal{L}_{P}^{(q)} / \mathcal{P})$\Comment{Update 3DGS}
    \State Sample $\boldsymbol{\eta} \in {N}(\boldsymbol{0}, \boldsymbol{1})$ \Comment{Exploration noise}
    \State $\hat{\boldsymbol{\mathrm{v}}}_{\tau+1} = \boldsymbol{\mathrm{v}}_{\tau} + a\nabla_{\boldsymbol{\mathrm{v}}_{\tau}}\log \mathcal{P} + b \boldsymbol{\eta}$ \Comment{LMC step}
    \State $q =\operatorname*{argmin}_{j \in {V}_{\text{vrt}}} |\hat{\boldsymbol{\mathrm{v}}}_{\tau+1} - \boldsymbol{\mathrm{v}}_{j}|$ \Comment{Nearest view}
    \State Update $\boldsymbol{\mathrm{v}}_{\tau+1} = \boldsymbol{\mathrm{v}}_{q}$
\EndFor
\State \Return $\{\mathcal{G}_i\}$
\end{algorithmic}
\end{algorithm}
\vspace{-46pt}
\end{minipage}
\end{wrapfigure}
To achieve this, we model the log-likelihood of the posterior distribution $\mathcal{P}$ as the L1 distance between the 3DGS rendering and the generated appearance, and parameterize the pose vector $\boldsymbol{\mathrm{v}} \in \mathbb{R}^{7}$ using the {camera center and quaternion}, which is steered by the gradient of the posterior distribution $\mathcal{P}$ and a stochastic exploration $\boldsymbol{\eta}$. We quantize the continuous update of $\boldsymbol{\mathrm{v}}$ at each LMC step to be the pose vector of the closest virtual viewpoint in ${V}_{\text{vrt}}$, so as to effectively constrain the sampling domain and enable efficient pre-caching of the generated images. Note that the LMC sampling only depends on the local gradient of $\mathcal{P}$, and hence induces minimal overhead leveraging the fast backward pass of Gaussian splatting. The proposed view-adaptive sampling is described in Algorithm~\ref{algo:mcmc} and an example of view-adaptive sampling update has been provided by the blue arrows in Figure \ref{fig:PB&VS} (b).

\section{Experiments}
\label{sec:exp}

To demonstrate the effectiveness of InceptionGS, we conduct extensive experiments on challenging real-world large-scale scenes from GigaNVS~\cite{wang2024xscale} and small-scale scenes from MipNeRF360~\cite{barron2022mip}. We first describe the experimental setup for simulating unstructured view sampling and then compare InceptionGS against the state-of-the-art baseline approaches~\cite{liu20253dgs, chen2025mvsplat360, yu2024viewcrafter, liu2024novel, you2024nvs, wu2025difix3d+, bahmani2024ac3d} on the task of NVS. We also conduct ablation studies on the key components of our pipeline to validate the efficacy of each design choice, and perform additional sensitivity analyzes under varying severity levels to assess robustness.

\subsection{Evaluation protocols}
\label{subsec:protocol}
\noindent{\textbf{Dataset.}} The GigaNVS dataset~\cite{wang2024xscale} consists of seven large-scale sceneries, each containing thousands of high-quality, real-captured multi-view images. The unparalleled scene complexity, along with unstructured viewpoint and scale variations, reveals a high-degree of non-uniformity in view density, making it a suitable testbed for InceptionGS. We also conduct experiments on all publicly available scenes from the MipNeRF360 dataset~\cite{barron2022mip} to demonstrate the broad applicability and robustness of our approach beyond large-scale outdoor settings.

\noindent{\textbf{Setup.}} To enable quantitative evaluation, we simulate unstructured view sampling by selectively holding out real-captured images as ground truth for NVS. Specifically, we cluster all viewpoints into six disjoint groups using farthest point sampling based on a hybrid distance metric considering both camera position and orientation. Starting from a random seed, each new cluster centroid is selected to maximize its minimal distance to the previously chosen ones, and the remaining viewpoints are assigned to the nearest centroid. We then randomly select one cluster and hold out 90\% of its viewpoints for evaluation, with the centroid of this cluster serving as the target view. During training, only the pose of target view is used as an anchor for pseudo-view generation, indicating the worst reconstructed region. During evaluation, we assess the bootstrapping effect using the images within the target-view cluster as ground truth rather than evaluating only the target view itself. We also uniformly hold out 5\% of viewpoints from all other groups to evaluate the quality of well-covered regions, in line with conventional NVS.

\noindent{\textbf{Baselines.}} We first compare against camera-controlled image-to-video(I2V) diffusion approaches, including MVSplat360~\cite{chen2025mvsplat360}, GEN3C~\cite{ren2025gen3c}, NVS-Solver~\cite{you2024nvs}, ViewCrafter~\cite{yu2024viewcrafter} and AC3D~\cite{bahmani2024ac3d}, multi-view diffusion approach Stable Virtual Camera(SEVA)\cite{zhou2025stable} and See3D~\cite{Ma2025See3D}, as well as the single-step scene-inpainting approach Difix3D+~\cite{wu2025difix3d+}, in terms of generative repairing for problematic regions. To adapt these methods to the unstructured sampling setting, we make minimal yet necessary modifications, running video diffusion strictly on the same set of candidate virtual views as ours. After obtaining the generated images at all virtual views, we optimize a 3DGS per-scene taking as inputs all real-captured training views and generated virtual views. The quantitative evaluation is performed on the final 3DGS renderings. We also compare against per-scene optimization based 3DGS extensions, including PGSR~\cite{chen2024pgsr}, 3DGS-MCMC~\cite{kheradmand20243d}, SplatFormer~\cite{chen2024splatformer} and DropGaussian~\cite{park2025dropgaussian}, using all real-captured training views. Please refer to the Appendix for more details. 

\noindent{\textbf{Metrics.}} We use PSNR, LPIPS, and Frechet Inception Distance (FID)~\cite{heusel2017gans} to evaluate reconstruction precision, visual fidelity, and generation quality.

\subsection{Comparative results}
\label{subsec:benchmark}
In the following, we present both quantitative and qualitative results, where the proposed method consistently outperforms state-of-the-art baselines.
 
\begin{wraptable}{r}{0.5\linewidth}
\vspace{-30pt}
\caption{Quantitative comparisons on the {GigaNVS}~\cite{wang2024xscale} and MipNeRF360~\cite{barron2022mip} dataset under unstructured view sampling.}
\label{tab:quantitative}
\centering
\resizebox{\linewidth}{!}{
\newcommand{\tabincell}[2]{\begin{tabular}{@{}#1@{}}#2\end{tabular}}
\footnotesize
\setlength{\tabcolsep}{0.3mm}{
\begin{tabular}{c|ccc|ccc}
\toprule
\multirow{2}{*}{\tabincell{c}{}} & \multicolumn{3}{c|}{GigaNVS~\cite{wang2024xscale}} & \multicolumn{3}{c}{MipNeRF360~\cite{barron2022mip}} \\ 
{} & PSNR$\uparrow$ & LPIPS$\downarrow$ & FID$\downarrow$ & PSNR$\uparrow$ & LPIPS$\downarrow$ & FID$\downarrow$\\ \midrule
MVSplat360~\cite{chen2025mvsplat360} & 13.59 & 0.602 & 168.62 & 13.94 & 0.547 & 180.74 \\
ViewCrafter~\cite{yu2024viewcrafter} & 15.01 & 0.565 & 138.39 & 16.18 & 0.544 & 171.71 \\
NVS-Solver~\cite{you2024nvs} & 15.54 & 0.530 & 128.47 & 16.90 & 0.546 & 161.20 \\
AC3D~\cite{bahmani2024ac3d} & 14.67 & 0.638 & 177.49 & 15.72 & 0.641 & 161.37 \\
Difix3D+~\cite{wu2025difix3d+} & 15.86 & 0.418 & 71.39 & 19.08 & 0.321 & 66.73 \\
SEVA~\cite{zhou2025stable} & 15.37 & 0.563 & 135.48 & 16.83 & 0.520 & 186.52 \\
GEN3C~\cite{ren2025gen3c} & 15.24 & 0.517 & 122.97 & 16.72 & 0.473 & 154.57 \\
See3D~\cite{Ma2025See3D} & 15.93 & 0.488 & 103.89 & 17.92 & 0.437 & 122.02 \\
SplatFormer~\cite{chen2024splatformer} & 14.28 & 0.511 & 128.13 & 15.88 & 0.402 & 125.54 \\
PGSR~\cite{chen2024pgsr} & 15.96 & 0.442 & 91.27 & 18.85 & 0.328 & 87.79 \\
3DGS-MCMC~\cite{kheradmand20243d} & 16.54 & 0.438 & 109.92 & 18.92 & 0.315 & 83.59 \\
DropGaussian~\cite{park2025dropgaussian} & 16.42 & 0.562 & 141.74 & 19.42 & 0.356 & 87.82 \\
\midrule
Ours (w/o FT) & 14.97 & 0.610 & 176.40 & 17.09 & 0.470 & 164.67 \\
Ours (w/o PB) & 16.97 & 0.376 & 73.35 & 20.48 & 0.301 & 64.89 \\ 
Ours (w/o VS) & 16.88 & 0.394 & 72.89 & 20.28 & 0.312 & 66.59 \\
\textbf{Ours (Full)} & \textbf{17.24} & \textbf{0.360} & \textbf{64.42} & \textbf{20.72} & \textbf{0.297} & \textbf{62.78} \\ 
\bottomrule
\end{tabular}
}
}
\vspace{-20pt}
\end{wraptable} 
 \noindent{\textbf{Quantitative results.}} In Table~\ref{tab:quantitative}, we quantitatively compare against the baselines and report the mean metrics across all test views and scenes. Our method outperforms all compared baselines by a noticeable margin and, remarkably, achieves a {10\%} reduction in average FID and a {14\%} reduction in average LPIPS relative to the best-performing diffusion method~\cite{wu2025difix3d+}, underscoring the superiority of InceptionGS in bootstrapping problematic regions of large-scale scenes. Note that diffusion based methods have difficulty generalizing to unstructured view sampling, primarily due to scene complexity, dramatic variations in pose and scale, and lack of global context in the reference image.

\begin{figure*}[htbp]
    \vspace{-15pt}
    \centering
    \newcommand{\colw}{0.19}
    \newcommand{\figw}{1.0} 
    \includegraphics[width=\figw\textwidth,trim={0cm 0cm 0cm 0cm},clip]{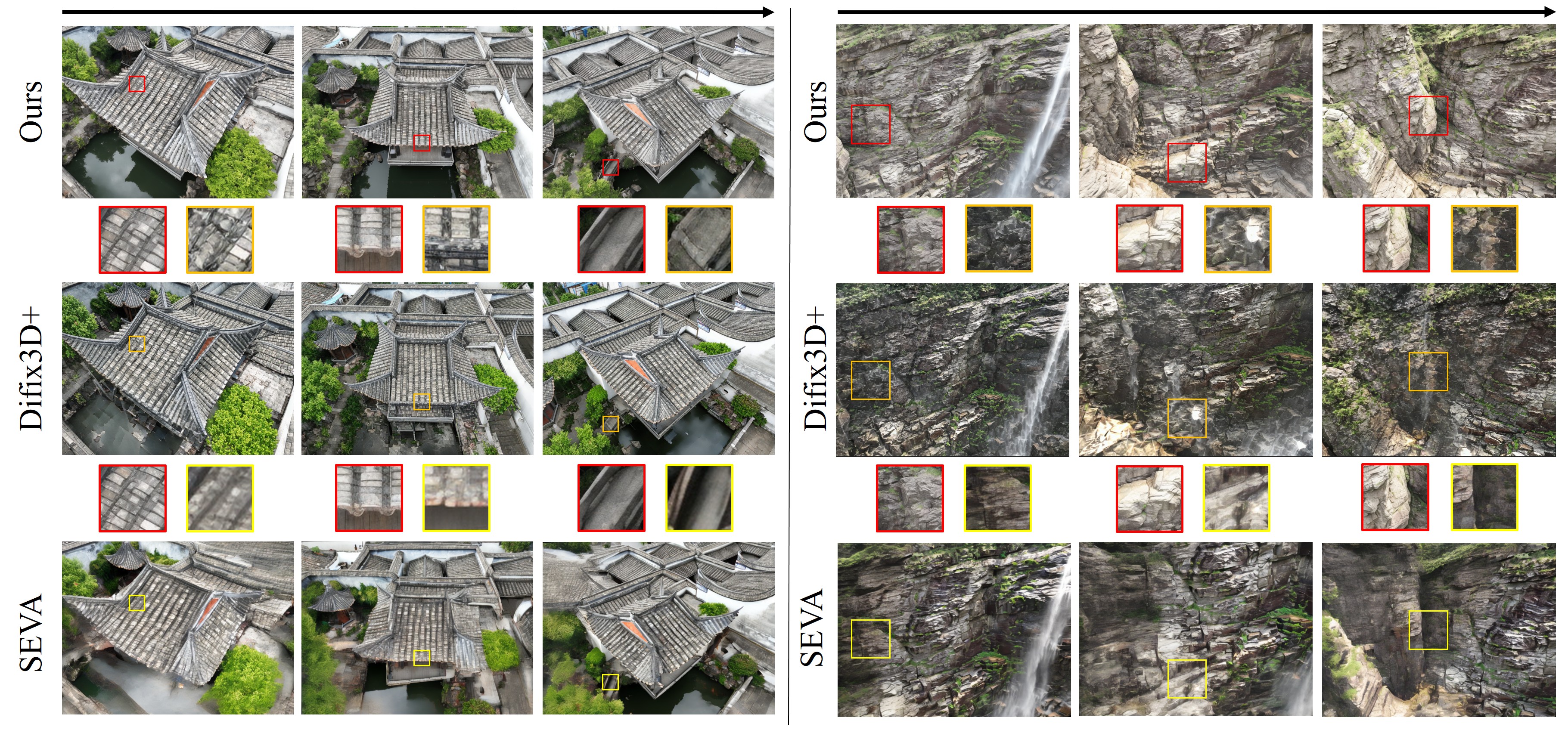}
    \hfill
\vspace{-20pt}
\caption{Visual comparisons on camera-controlled generation against Difix3D+~\cite{wu2025difix3d+} and SEVA~\cite{zhou2025stable}. Our scene-adaptive, reconstruction-guided LDM enables significantly superior visual quality, 3D consistency, and controllability.}
\vspace{-20pt}
\label{fig:gen_compare}
\end{figure*}
\begin{figure*}[!htbp]
    \centering
    \newcommand{\colw}{0.19}
    \newcommand{\figw}{1} 
    \includegraphics[width=\figw\textwidth,trim={0cm 0cm 0cm 0cm},clip]{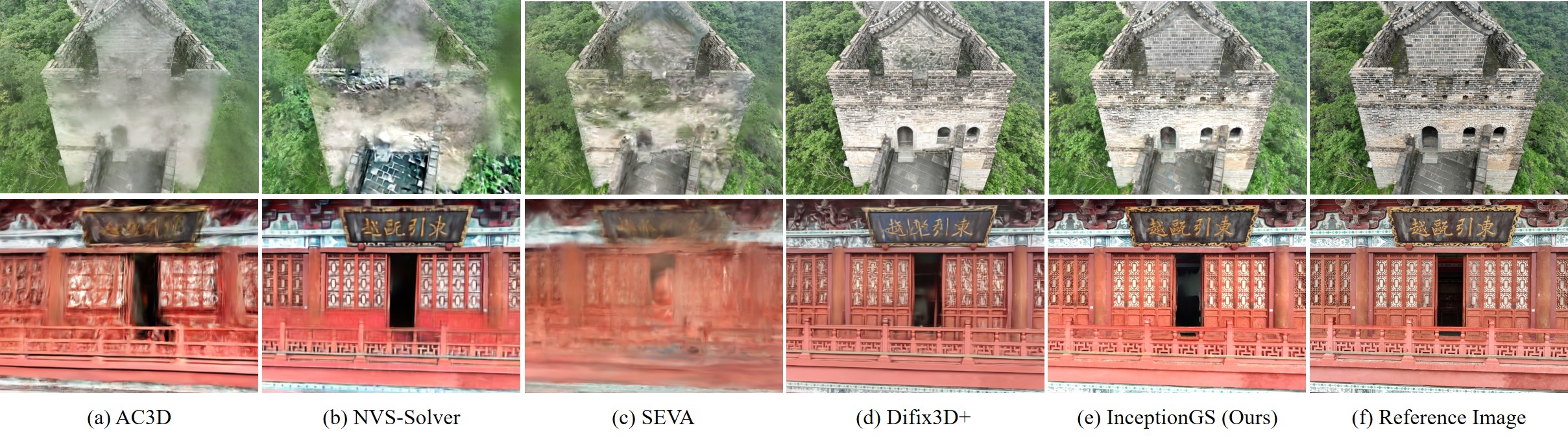}
    \hfill
\vspace{-0.7cm}
\caption{Qualitative 3DGS comparisons with generative approaches~\cite{bahmani2024ac3d, zhou2025stable, you2024nvs, wu2025difix3d+} on GigaNVS~\cite{wang2024xscale}.}
\label{fig:compare}
\end{figure*}
\begin{figure*}[!htbp]
    \centering
    \newcommand{\colw}{0.19}
    \newcommand{\figw}{1.0} 
    \includegraphics[width=\figw\textwidth,trim={0cm 0cm 0cm 0cm},clip]{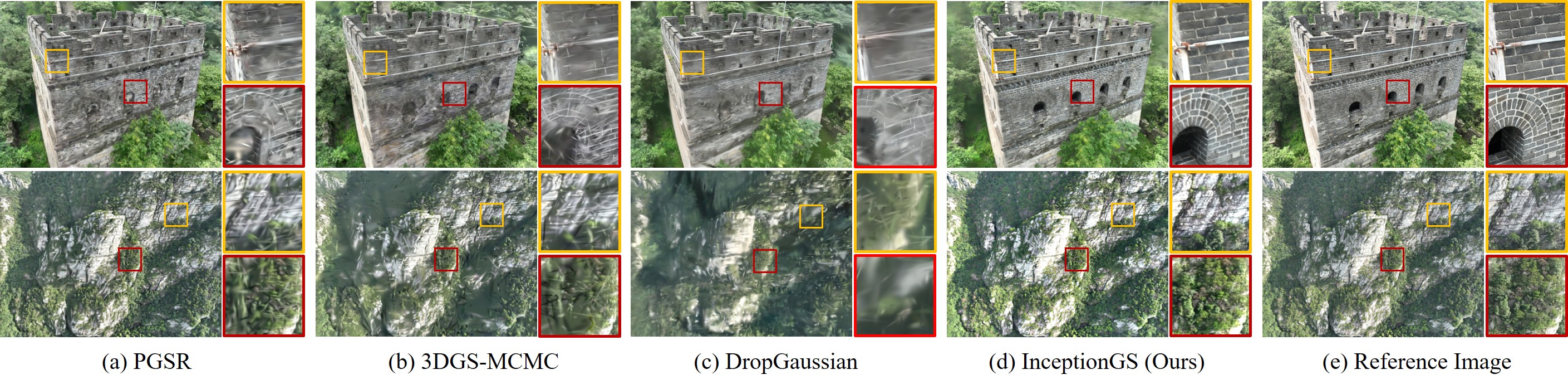}
    \hfill
\vspace{-0.7cm}
\caption{Visual demonstrations on the effectiveness of InceptionGS. 
}
\vspace{-15pt}
\label{fig:gs_compare}
\end{figure*}
\noindent{\textbf{Qualitative results.}} In Figure~\ref{fig:gen_compare}, we visualize the raw generation results from our scene-adaptive LDM and the top-performing diffusion-based methods~\cite{wu2025difix3d+, zhou2025stable}. Existing generation approaches are prone to structural or appearance hallucinations in complex large-scale scenes, especially when the camera deviates significantly from the conditioning view. By contrast, our method directly learns the scene-specific geometry-appearance correspondence, enabling 3D consistent, faithful generation grounded on reconstruction, while robustly handling drastic camera movements and scale variations. The visual comparisons of the lifted 3DGS fields are presented in Figure~\ref{fig:compare}. Notably, all diffusion based baselines exhibit excessive blurriness, structural distortions, or floating artifacts due to multi-view inconsistent and deteriorated generation results. On the contrary, InceptionGS enables visually pleasing, detail-preserving bootstrapping by softly incorporating the scene-adaptive generative prior. Please refer to the Appendix for more qualitative results.

\FloatBarrier

\subsection{Ablation study}
\label{subsec:ablation}
\noindent{\textbf{Effectiveness of InceptionGS.}} The quantitative ablations for the full generative bootstrapping pipeline is reported in Table~\ref{tab:quantitative} (\textit{Ours (Full)} V.S. \textit{PGSR}). Remarkably, InceptionGS improves the overall visual quality, reducing the average FID by 32\%. The visual comparisons are presented in Figure~\ref{fig:gs_compare}. By applying InceptionGS, the problematic regions are effectively repaired with fine details, demonstrating its capability to handle unstructured view sampling problem without the need for additional labor-intensive real-world data collection. 

\noindent{\textbf{Ablations on key components.}} In Table~\ref{tab:quantitative}, we also conduct ablations on proposed photometric blending (PB), view-adaptive sampling (VS), and fine-tuning (FT) of the generic generative model. All three components are crucial in boosting performance -- while photometric blending can utilize reliable image warping wherever possible to further improve textural details, view-adaptive sampling mitigates the conflicts between reconstructive and generative supervisions and preserves the quality of well-reconstructed regions. Moreover, our adapted LDM, trained with ControlNet-based geometric conditioning, effectively captures scene-specific statistics to enable controllable and spatially consistent novel view synthesis. Note that the primary gain of InceptionGS comes from the scene-adaptive generative model, given the remarkable performance gap between both ablative versions (\textit{Ours (w/o PB)} and \textit{Ours (w/o VS)}) and the baselines.


\begin{wraptable}{r}{0.5\linewidth}
\vspace{-25pt}  
\centering

\caption{Ablation on increased scarcity.}
\label{tab:scarcity}
\scriptsize
\begin{tabular}{cccccc}
\hline
\multicolumn{1}{c}{\#visible views} & PSNR$\uparrow$ & SSIM$\uparrow$ & LPIPS$\downarrow$ & FID$\downarrow$\\ \hline
4 (Original setup) & \textbf{26.82} & \textbf{0.822} & \textbf{0.231} & \textbf{50.89} \\ 
2 (More holes) & 26.60	& 0.821	& 0.234	& 53.35 \\ 
\hline
\end{tabular}

\vspace{-0.2em}

\caption{Ablation on the number of virtual views injected.}
\label{tab:view-number}
\scriptsize
\begin{tabular}{cccccc}
\hline
\multicolumn{1}{c}{\#virtual views} & PSNR$\uparrow$ & SSIM$\uparrow$ & LPIPS$\downarrow$ & FID$\downarrow$\\ \hline
1.5$\times$ & \textbf{16.30} & 0.551 & 0.383 & 54.12 \\ 
1$\times$ & 16.28 & \textbf{0.560} & \textbf{0.377} & 53.79 \\ 
0.5$\times$ & 16.24	& 0.541	& 0.389	& \textbf{53.71} \\ 
\hline
\end{tabular}

\vspace{-20pt} 
\end{wraptable}
\noindent{\textbf{Sensitivity to severity.}} 
We further evaluate the robustness of the proposed method to varied levels of severity.
Specifically, to simulate a more challenging reconstruction setting, we reduce by 50\% the number of input views that observe the held-out region, thereby significantly limiting geometric and photometric cues. As reported in Table~\ref{tab:scarcity} on the Room scene from MipNeRF360 dataset, the reconstruction quality exhibits only a marginal degradation. The results indicate that our method is robust to different levels of view scarcity and geometry quality. This robustness mainly arises from training ControlNet on inputs rendered on-the-fly from a continuously updated GS model, allowing it to generalize to varied severity effectively.
In addition, we adjust the number of virtual viewpoints by 50\% to investigate the sensitivity of our framework to different levels of generative guidance, with the results on the Lanes-and-Alleys scene from GigaNVS dataset shown in Table~\ref{tab:view-number}. Interestingly, the performance remains quite stable across different configurations. Such insensitivity indicates that our framework does not rely on a precise tuning of virtual view quantity. The soft injection strategy distributes generative supervision adaptively in the view space, ensuring that additional virtual views do not introduce redundancy, while fewer views still provide sufficient guidance.
Across both settings, our method maintains stable performance, which indicates our robustness to varied levels of severity. 

\noindent{\textbf{Demonstrations on iterative bootstrapping.}} In Table~\ref{tab:iterative}, we conduct experiments by iterating our bootstrapping pipeline as mentioned in the beginning of Section~\ref{sec:method}, where a second round is performed taking both the original images and the repaired rendered images from the first round as input. We investigate two different scenarios for the target views: (1) identical target views across both rounds ($j_{(1)}^{\star} =  j_{(2)}^{\star} $), and (2) different target views oriented toward separate underobserved regions ($j_{(1)}^{\star} \neq  j_{(2)}^{\star} $). For both scenarios, the evaluation is performed on the same 122 test views comprising views focused on underobserved regions and views uniformly sampled across the entire scene. The metrics are reported on the TW-Pavilion (Day) scene from the GigaNVS dataset.


\begin{table}[htbp]
\vspace{-10pt}
\caption{Effects of iterative bootstrapping.}
\label{tab:iterative}
\centering
\footnotesize
\setlength{\tabcolsep}{0.8 mm}{
\begin{tabular}{l|cccc}
\toprule
{} & PSNR $\uparrow$ & SSIM $\uparrow$ & LPIPS $\downarrow$ & FID $\downarrow$ \\ \midrule
1st round with $j_{(1)}^{\star}$ &14.53	& 0.447 & 0.508 & 143.93 \\ 
2nd round with $j_{(2)}^{\star} = j_{(1)}^{\star} $ & 14.66 & 0.448 & 0.514 & 140.79 \\ 
2nd round with $j_{(2)}^{\star} \neq j_{(1)}^{\star}$ & \textbf{16.28} & \textbf{0.551} & \textbf{0.377} & \textbf{53.79} \\ \bottomrule
\end{tabular}
}
\vspace{-25pt}
\end{table}

\begin{figure*}[htbp]
    \centering
    \newcommand{\colw}{0.19}
    \newcommand{\figw}{1} 
    \includegraphics[width=\figw\textwidth,trim={0cm 0cm 0cm 0cm},clip]{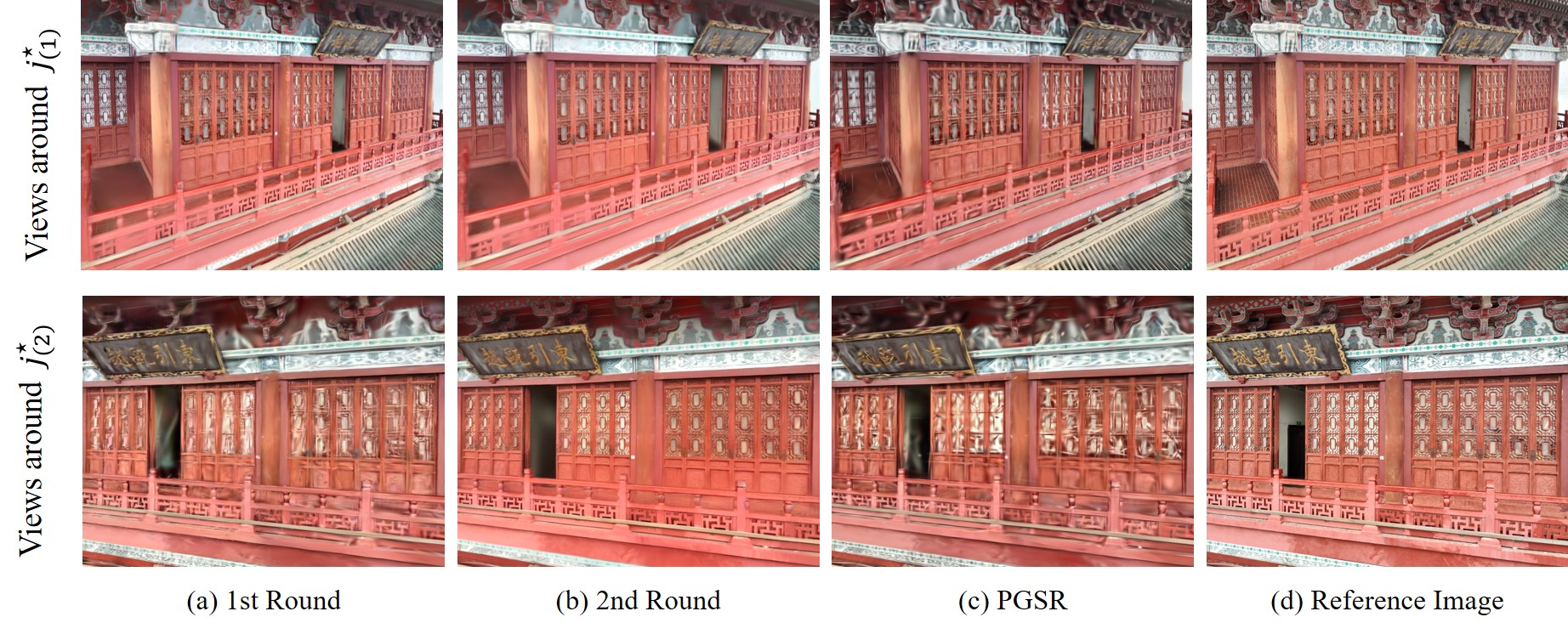}
    \hfill
\vspace{-0.7cm}
\caption{Qualitative demonstrations of iterative bootstrapping on  different views targeting separate underobserved regions ($j_{(1)}^{\star} \neq  j_{(2)}^{\star} $).}
\vspace{-10pt}
\label{fig:iter}
\end{figure*}

According to Table~\ref{tab:iterative}, the results appear to be stable when identical target views are used for two consecutive rounds. This indicates the effectiveness of our soft blending and sampling strategy, where a single round is generally enough for repairing a specific region and further rounds will not bring negative impacts on reconstruction. When different target views are specified, as shown in Figure~\ref{fig:iter}, the results improve after a second round of repairing without deteriorating the quality of previously well-reconstructed areas, thus demonstrating the efficacy of iterative bootstrapping. 
Due to space limit, more ablation studies can be found in the Appendix.
\section{Conclusion}
\label{sec:conclusion}
We introduce InceptionGS, a novel approach to address the unstructured view sampling challenge related to real-world large-scale scenes. By adapting the generic generative prior with scene-specific geometry-appearance correspondence, InceptionGS can incorporate this reliable prior in a view-adaptive manner and incrementally refine the 3DGS field without requiring additional data collection. Extensive experiments on challenging benchmarks demonstrate the significant superiority of InceptionGS, highlighting its potential in advancing truly immersive large-scale scene digitization.

\noindent{\textbf{Limitation \& future work.}} Despite the compelling results, stage I takes approximately 30 minutes with peak VRAM of 30GB, while stage II takes around 25 minutes with peak VRAM below 15GB. For future work, we aim to 
explore more advanced optimization scheme like test-time training to reduce per-scene overhead and improve scalability. 

\noindent{\textbf{Acknowledgements.}} This work is supported in part by Natural Science Foundation of China (NSFC) under contract No. 62125106 and 62427804, in part by Tsinghua University Dushi Program (No.20251080107), in part by the Beijing Outstanding Young Scientist Program under contract No. JWZQ20240101009, in part by the XPLORER PRIZE.

\clearpage  


%
%
\bibliographystyle{splncs04}
\bibliography{main}
\end{document}